\documentclass{article}

\usepackage{PRIMEarxiv}

\usepackage[utf8]{inputenc} 
\usepackage[T1]{fontenc}    
\usepackage{hyperref}       
\usepackage{url}            
\usepackage{booktabs}       
\usepackage{amsfonts}       
\usepackage{nicefrac}       
\usepackage{microtype}      
\usepackage{lipsum}
\usepackage{fancyhdr}       
\usepackage{graphicx}       
\graphicspath{{media/}}     

\usepackage{import}
\usepackage{booktabs}
\usepackage{multicol}
\usepackage{multirow}

\usepackage{amsmath,amsfonts,bm}

\def\eqref#1{equation~\ref{#1}}

\def\1{\bm{1}}

\DeclareMathAlphabet{\mathsfit}{\encodingdefault}{\sfdefault}{m}{sl}
\SetMathAlphabet{\mathsfit}{bold}{\encodingdefault}{\sfdefault}{bx}{n}

\usepackage{amsmath}
\usepackage{amssymb}

\usepackage[utf8]{inputenc} 
\usepackage[T1]{fontenc}    
\usepackage{hyperref}       
\usepackage{url}            
\usepackage{booktabs}       
\usepackage{amsfonts}       
\usepackage{nicefrac}       
\usepackage{microtype}      
\usepackage{xcolor}         
\usepackage{natbib}
\PassOptionsToPackage{numbers,sort,comma,square}{natbib}
\usepackage{amsmath}
\usepackage{graphicx}
\usepackage{subcaption}
\usepackage{enumitem}
\usepackage[table]{xcolor}
\usepackage{arydshln}

\newcommand{\MBPO}{\texttt{MBPO}}

\title{Modality-Balancing Preference Optimization of Large Multimodal Models by Adversarial Negative Mining
}

\author{
  Chenxi Liu, Tianyi Xiong, Yanshuo Chen, Ruibo Chen, Yihan Wu, Junfeng Guo, Tianyi Zhou, Heng Huang \\
  University of Maryland, College Park \\
  \texttt{cxliu539@umd.edu} \\
}

\begin{document}
\maketitle

\begin{abstract}
The task adaptation and alignment of Large Multimodal Models (LMMs) have been significantly advanced by instruction tuning and further strengthened by recent preference optimization. Yet, most LMMs still suffer from severe modality imbalance during reasoning, i.e., outweighing language prior biases over visual inputs, which bottlenecks their generalization to downstream tasks and causes hallucinations. 
However, existing preference optimization approaches for LMMs do not focus on restraining the internal biases of their Large Language Model (LLM) backbones when curating the training data. 
Moreover, they heavily rely on offline data and lack the capacity to explore diverse responses adaptive to dynamic distributional shifts during training. Meanwhile, Group Relative Policy Optimization (GRPO), a recent method using online-generated data and verified rewards to improve reasoning capabilities, remains largely underexplored in LMM alignment. 
In this paper, we propose a novel preference learning framework, Modality-Balancing Preference Optimization (\MBPO{}), to address the modality imbalance in LMMs. \MBPO{} constructs a more effective offline preference dataset by generating hard negatives, i.e., rejected responses misled by LLM biases due to limited usage of visual information, through adversarial perturbation of input images.  
Moreover, \MBPO{} leverages the easy-to-verify nature of close-ended tasks to generate online responses with verified rewards. GRPO is then employed to train the model with offline-online hybrid data. Extensive experiments demonstrate that \MBPO{} can enhance LMM performance on challenging vision-language tasks and effectively reduce hallucinations. Source code is available at \url{https://github.com/DawnLIU35/MBPO}
\end{abstract}

\section{Introduction}
\label{introduction}

\emph{Large Multimodal Models} (LMMs) have achieved incredible success by integrating vision models with pre-trained \emph{Large Language Models} (LLMs) through instruction tuning, enabling effective adaptation to diverse visual tasks~\citep{liu2023visual,liu2024llavanext,Qwen2.5-VL,tong2024cambrian,chen2024internvl,2023GPT4VisionSC,xiong2024llavacritic,chen2023gpt,chen2024your,liu2024few}. Despite their strong performance across complex visual understanding scenarios, LMMs still face several fundamental challenges: achieving proper alignment between multimodal inputs~\citep{li2024llava,liu2023visual}; collecting and effectively leveraging high-quality aligned multimodal data with accurate annotations~\citep{tong2024cambrian,luo2024mmevol}; and mitigating hallucination, where models generate content disconnected from or contradicting the visual evidence~\citep{yu2024rlhf,zhao2023beyond}. Furthermore, recent studies show that LMMs suffer from the modality imbalance problem, tending to over-rely on their language backbone while underutilizing the rich information available in visual inputs~\citep{liu2024paying,jiang2024modality}, thus leading to problematic behaviors such as incorrect visual perception and hallucinated responses.

To further improve task adaptation and alignment with human intent, recent studies~\citep{yu2024rlhf,zhou2024calibrated,lu2025damo} adopt preference learning as a post-training strategy for LMMs, enhancing performance in general vision-language tasks and reducing hallucination. 
Benefiting from the simplified reward parameterization introduced by \emph{Direct Preference Optimization} (DPO)~\citep{rafailov2023direct}, some works~\citep{yu2024rlhf,pi2024strengthening,cui2024fine,jiang2024modality,yu2024rlaifv, amirloo2024understanding} propose various strategies for constructing pairwise preference datasets, typically selecting high-quality responses as preferred examples and hallucinated ones as rejected.
While these methods help align model outputs with human preferences, they do not explicitly tackle the modality imbalance issue—where LMMs tend to over-rely on the linguistic priors of the language backbone rather than grounding their predictions in visual input.
Furthermore, the inherently offline nature of DPO—relying exclusively on pre-collected model responses—limits its ability to adapt to distributional shifts during training, thereby hindering optimization effectiveness~\citep{chen2024self,chen2024optune}.

In contrast, the recently proposed \emph{Group Relative Policy Optimization} (GRPO)~\citep{shao2024deepseekmath} improves reasoning capabilities by utilizing online model-generated trajectories with verifiable reward signals~\citep{guo2025deepseek}. Recent studies~\citep{chen2025r1v,shen2025vlm,zheng2025easyr1} have explored the potential of using GRPO to visual reasoning tasks, such as multimodal math problems and visual perception. However, the broader potential of reinforcement learning with verified rewards for general multimodal alignment remains largely underexplored.

In this paper, we propose Modality-Balancing Preference Optimization (\MBPO{}), a novel framework that combines both offline and online preference data to address modality imbalance and improve alignment in LMMs. This framework comprises two complementary components: (1) an offline pairwise preference dataset constructed using adversarially mined negative responses, and (2) an online dataset with verifiable rewards collected dynamically during training.
\begin{itemize}[leftmargin=2em]
    \setlength{\itemsep}{0.3em}
    \item For the offline dataset, we focus on addressing modality imbalance issue, where the model tends to rely more on the language backbone’s prior knowledge than on visual evidence. We first introduce an image information gain metric that quantifies how much visual content is utilized in a response. To generate rejected responses with low image information gain and high modality imbalance, we apply adversarial perturbations to the input image to reduce the model’s confidence in the original ground-truth response. The perturbed image is then used, together with the original instruction, to produce a less visually grounded rejected response.
    \item For the online dataset, we leverage closed-ended visual instruction-tuning data (i.e., multiple-choice and yes/no questions) with verifiable answers. During training, the model generates multiple candidate responses for each input instruction, and rewards are assigned based on factual correctness. To avoid generating extremely short responses, we add a simple prompt instruction and an extra format reward to the online dataset. By adapting to distributional shifts throughout training, these reward signals enable more effective model alignment. 
\end{itemize}
 We jointly optimize the model using both offline and online data through the Group Relative Policy Optimization (GRPO) objective. Experimental results on a wide range of vision language tasks and hallucination benchmarks demonstrate that \MBPO{} significantly mitigates modality imbalance and enhances overall performance.

Overall, \textbf{our contributions} can be summarized as follows:
\begin{itemize}[leftmargin=2em]
    \setlength{\itemsep}{0.3em}
    \item We propose \MBPO{}, a novel framework that addresses modality imbalance in large multimodal models (LMMs) to improve alignment. By mining adversarial images to construct rejected responses, \MBPO{} explicitly incentivizes LMMs to incorporate visual information during response generation.

    \item We leverage the easy-to-verify nature of close-ended data as an online dataset and use a simple prompt instruction along with a format reward to encourage the model to generate more diverse responses, including verifiable single-word answers and corresponding explanations.

    \item Experiments across general vision-language tasks and hallucination benchmarks demonstrate that \MBPO{} effectively enhance LMM performance while effectively mitigating modality imbalance.
\end{itemize}

\section{Related Work}
\label{related_work}

\textbf{Multimodal Preference Learning}. Preference learning is a proven method to align pretrained LLMs~\citep{ouyang2022training,mcaleese2024llm} and LMMs~\citep{sun2023aligning} with human intentions and reduce model hallucination. Specifically, \emph{Direct Preference Optimization} (DPO)~\citep{rafailov2023direct} has been widely adopted for its elimination of an explicit reward model, enabling direct optimization over pairs of preferred and rejected responses. Prior works have collected multimodal preference datasets using human annotations~\citep{yu2024rlhf} or AI-generated feedback~\citep{li2023silkie,xiong2024llavacritic}.  Another line of papers focus on self-rewarding~\citep{yuan2024selfrewarding,chen2024self} mechanisms, gathering preference data from model-generated response without external supervision. These approaches typically involve the design of evaluation prompts~\citep{wang2024enhancing}, sentence-level search strategies~\citep{zhou2024calibrated} or decomposition into fine-grained judgments~\citep{yu2024rlaifv,cui2024fine}.
Although some methods re-collect preference data for multi-round iterative training, the inherently offline nature of DPO leads them to rely heavily on pre-collected model responses within each epoch, making it difficult to adapt to distribution shifts during training. In contrast, our method combines online and offline samples for both dynamic and consistent preference alignment.

\textbf{Noise Injection in Multimodal Preference Learning} While human annotations are costly and AI-generated feedback is susceptible to reward hacking~\citep{skalse2022defining} and lacks verifiability, some studies create rejected responses by deliberate error injections. some works~\citep{pi2024strengthening,zhou2024aligning} apply Gaussian distortions to input images and employ LLM or LMM to introduce hallucinated responses, while \cite{wang2024mdpo} apply random cropping on images. More recently, \cite{liu2025noisyrollout} use distorted image inputs in GRPO training to enhance LMM reasoning in multimodal math. However, rejected responses generated with random image distortion or external rewriting may not yield clearly incorrect outputs and often lie far from the model generation distribution. Our work focuses on adversarial inputs that produce in-domain, instruction-following responses that are incorrect yet highly probable under the model’s distribution.

\textbf{Multimodal RLVR.} Recent studies show that large-scale reinforcement learning significantly enhances LLM in complex reasoning~\citep{jaech2024openai,guo2025deepseek,team2025kimi}. Several concurrent works extend \emph{Reinforcement Learning with Verifiable Rewards} (RLVR), as used in Deepseek-R1 to multimodal settings. One line of research focuses on multimodal math~\citep{meng2025mm,huang2025vision}, academic questions~\citep{peng2025lmm,yang2025r1}, while others target visual perception tasks~\citep{yu2025perception} such as counting~\cite{chen2025r1v}, grounding~\citep{shen2025vlm}, detection~\citep{zhan2025vision}, and refering segmentation~\citep{liu2025seg}. In our paper, we extend RLVR to broader visual domains, including general visual question answering, open-ended visual chat and hallucination related tasks.

\section{Preliminaries}
\label{preliminaries}

\textbf{Adversarial Attacks} on images can mislead LMMs into generating incorrect or misleading responses. To expose worst-case vulnerabilities of the model, adversarial images can be crafted by \emph{Projected Gradient Descent} (PGD)~\citep{madry2017towards}, the multistep extension of the \emph{Fast Gradient Sign Method} (FGSM)~\citep{goodfellow2014explaining} that is widely regarded as the strongest first-order $\ell_\infty$ attack.  Beginning from either the clean input $x$ or a random point $x^{(0)} \!\sim\! \mathcal{U}\bigl(x-\epsilon,x+\epsilon\bigr)$ inside the $\ell_\infty$ ball of radius $\epsilon$, PGD perform $T$ iterative updates
\begin{equation}
x^{(t+1)} = \Pi_{B_\epsilon(x)}
\Bigl(
  x^{(t)} + \alpha \cdot \text{sign}\!
  \bigl(\nabla_{x} J(\theta,x^{(t)},y)\bigr)
\Bigr),
\qquad t = 0,\ldots,T-1,
\end{equation}
where $\alpha$ is the step size, $\theta$ is the parameter of model and $J(\theta,x,y)$ is the loss, and  
$\Pi_{B_\epsilon(x)}(\cdot)$ projects its argument back onto the $\ell_\infty$ ball
$B_\epsilon(x) = \{\tilde{x} : \lVert \tilde{x} - x \rVert_\infty \le \epsilon \}$.  
After the final iteration, PGD clips $x^{(T)}$ to the valid data range to obtain the adversarial example $x^{\text{adv}}$.  By following the steepest ascent direction at each step yet remaining within the prescribed perturbation budget, PGD yields perturbations that are imperceptible to humans but significantly degrade model performance, providing a stringent evaluation of robustness.

\textbf{Group Relative Policy Optimization (GRPO)}~\citep{shao2024deepseekmath,guo2025deepseek} has been proven effective on LLMs. Instead of relying on a critic model, which is typically as large as the policy model, this approach estimates the baseline using group scores. Specifically, for each question $q$, GRPO samples a set of outputs $\{o_1, o_2, \ldots, o_G\}$ from the old policy $\pi_{\theta_{\text{old}}}$, and then updates the policy model $\pi_\theta$ by maximizing the following objective:

\begin{equation}
\begin{split}
\mathcal{J}_{GRPO}(\theta) 
    &= \mathbb{E}{[q \sim P(Q),\,\{o_i\}_{i=1}^G \sim \pi_{\theta_{\mathrm{old}}}(O\mid q)]} \\ 
    &\quad \frac{1}{G}\sum_{i=1}^G \Bigl\{
        \min\!\Bigl(\frac{\pi_\theta(o_i\mid q)}{\pi_{\theta_{\mathrm{old}}}(o_i\mid q)}A_i,\;\mathrm{clip}\Bigl(\frac{\pi_\theta(o_i\mid q)}{\pi_{\theta_{\mathrm{old}}}(o_i\mid q)},\,1-\epsilon,\,1+\epsilon\Bigr)\,A_i\Bigr) \\ 
    &\quad -\;\beta\;\mathbb{D}_{KL}\!\bigl(\pi_{\theta}\,\|\,\pi_{\mathrm{ref}}\bigr)
\Bigr\},
\end{split}
\label{eq:GRPO-obj}
\end{equation}

where $\epsilon$ and $\beta$ are hyperparameters, and $A_i$ denotes the advantage, which is computed based on a group of rewards $\{r_1, r_2, \ldots, r_G\}$ associated with the outputs in each group:

\begin{equation}
    A_i = \frac{r_i - {\text{mean}(\{r_1, r_2, \cdots, r_G\})}}{{\text{std}(\{r_1, r_2, \cdots, r_G\})}}.
\end{equation}

To prevent the updated policy $\pi_\theta$ from deviating too far from the stable reference $\pi_{\mathrm{ref}}$, GRPO loss has a \emph{Kullback-Leibler Divergence} term $\mathbb{D}_{KL}$ which is estimated with an unbiased estimator:

\begin{equation}
    \mathbb{D}_{KL}\left(\pi_{\theta} || \pi_{ref}\right) = \frac{\pi_{ref}(o_i|q)}{\pi_{\theta}(o_i|q)}- \log\frac{\pi_{ref}(o_i|q)}{\pi_{\theta}(o_i|q)} - 1,
\end{equation}
\section{Methodology}
\label{methodology}

\MBPO{} is a hybrid preference learning framework designed to enhance alignment and mitigate the modality imbalance problem in LMMs. It combines both offline and online preference data to provide stable yet adaptive reward signals throughout training. Section~\ref{sec:offline} introduces how \MBPO{} constructs the offline preference dataset, where the chosen responses are accurate and visually grounded, and the rejected responses rely heavily on the LLM backbone’s prior knowledge, neglecting visual information. These modality-imbalanced rejected responses are generated by adding adversarial noise to input images, which suppresses visual cues and triggers the prior biases from the LLM backbone. Section~\ref{sec:online} describes how \MBPO{} performs online exploration using closed-ended data with verifiable rewards. With a simple prompt instruction and an extra format reward, \MBPO{} enhances the model’s ability to explore diverse responses and dynamically adapt to distributional shifts during training.

\begin{figure}[t!]
  \centering
  \includegraphics[width=1\linewidth]{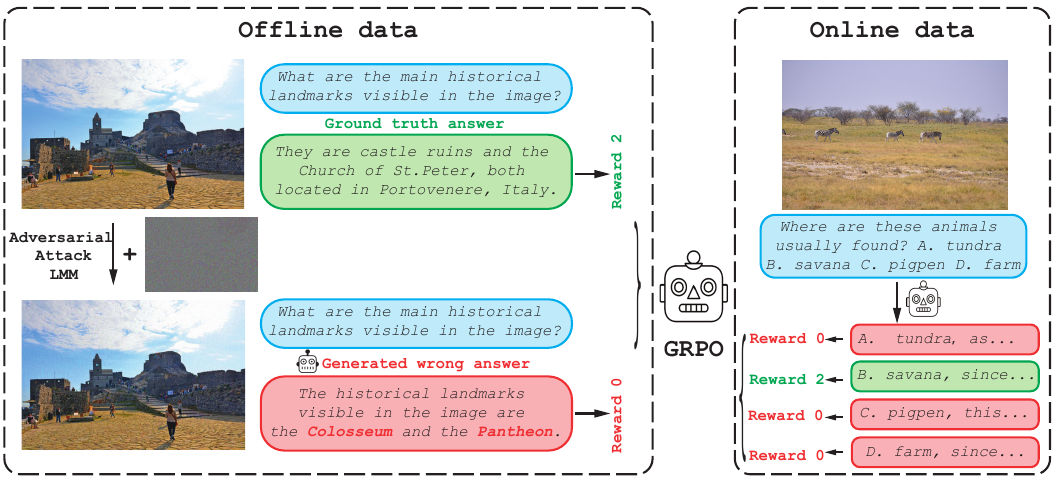}
  \caption{ 
  Overview of \MBPO{} framework. To construct the offline preference dataset, we generate adversarial perturbations for each input image to minimize the output probability of the chosen response. Rejected responses are then generated using these adversarially perturbed images. This process amplifies modality imbalance, causing the LMM to rely more heavily on the prior biases of its LLM backbone rather than the visual information. 
  In parallel, \MBPO{} incorporates an online dataset composed of closed-ended examples, where response correctness can be easily verified. During training, the LMM generates multiple responses, and verified rewards are assigned based on their correctness. Finally, the offline and online datasets are combined to optimize the LMM using the \MBPO{} loss in a hybrid training paradigm.
  }
  \label{fig:overview}
\end{figure}

An overview of our training pipeline is illustrated in Figure~\ref{fig:overview}.

\subsection{Offline Preference Data Construction}
\label{sec:offline}

Current LMMs often suffer from the modality imbalance problem that model responses overweigh the prior biases of the LLM backbone and underutilize the visual information from the image encoder, leading to incorrect or insufficient visual content in the output responses. To address this issue, \MBPO{} is designed to balance different input modalities to incorporate more accurate and relevant visual information into the the model responses. To quantify this, we propose a metric called \emph{Image Information Gain} (IIG), which measures the amount of visual information contained in the generated response. Given data consisting of a question $q$, an image $I$, and a response $o$, IIG is defined as:

\begin{equation}
\text{IIG}(o, q, I) = -\log p_{\theta}(o \mid q, I_b) + \log p_{\theta}(o \mid q, I)
\end{equation}

where $I_b$ denotes a blank image (all-zero pixels) of the same dimensions as $I$. This metric captures the difference in output probability when conditioned on the actual image versus a blank image with no information, using the same question and response. A larger IIG value indicates that the response $o$ incorporates more information from the image $I$. As the goal of \MBPO{} is to encourage LMMs to incorporate more visual information into their responses, we select data whose responses have high IIG scores from a visual instruction tuning dataset as our preference dataset and chosen responses.

The next step is to construct the corresponding rejected responses for the selected data. Compared to the chosen responses that contain rich image information, the rejected responses should include limited visual information and rely primarily on the prior biases of the LLM backbone. To generate the rejected responses, \MBPO{} adds adversarial noise to the image to minimize the output probability of the chosen responses:

\begin{equation}
I^{(t+1)} = \Pi_{B_\epsilon(\mathbf{I})} \left(
  I^{(t)} 
  + \alpha \cdot \text{sign} \left( 
    \nabla_{I} \left( -\log p_\theta(o_w \mid q, I) \right)
  \right)
\right),
\qquad t = 0, \ldots, T{-}1
\end{equation}

where $o_w$ is the winner/chosen response from the visual instruction dataset. $I^0$ is the original image from the visual instruction dataset, and we denote the final $I^T$ as the adversarial image $I_{\text{adv}}$. After obtaining the adversarial image, we sample a loser/rejected response using the same question $q$:
\begin{equation}
o_l \sim \pi_\theta(\cdot \mid q, I_{\text{adv}})
\end{equation}
As the adversarial image is perturbed to minimize the output probability of the correct chosen response, it loses visual information relevant to that response. When the model generates a new response using the adversarial image, it cannot effectively retrieve the visual information from the image and instead relies on the prior biases of the LLM backbone.

The chosen responses from the visual instruction dataset and the generated rejected responses using adversarial images constitute our offline preference dataset: $D_{\text{offline}} = \left\{ \left(q, I, o_w, o_l \right) \right\}$. During our training, we assign hard rewards to the offline data. Specifically, a reward of $2$ is given to the chosen response, and a reward of $0$ is assigned to the rejected response. 

\subsection{Online Preference Data}
\label{sec:online}
Although training on offline datasets can improve a model's performance, they still face several limitations. First, they cannot adapt to the latest distribution shifts during training, limiting their training effectiveness on the offline data~\citep{chen2024self}. Moreover, offline data typically consists of pairwise preference annotations, which represent only a limited set of possible model responses. In contrast, online preference learning methods~\citep{peng2025lmm,guo2025deepseek} generate multiple responses using the latest model weights, allowing optimization over the current output distribution and enabling the sampling of more possible responses. Furthermore, they can provide accurate feedback to online generations using verified rewards, rather than relying on unreliable reward models or costly human verification.

In the visual instruction dataset, we observe that the responses for multiple-choice and yes/no questions are easy to verify using verifiable checking~\citep{shao2024deepseekmath}. Therefore, we construct our online preference data using all the multiple-choice and yes/no samples from \texttt{MMSeed}, totaling around $2k$ examples. 
However, in the original visual instruction dataset, the multiple-choice data are prompted with \texttt{"Answer with the option's letter from the given choices directly."}, which results in responses with limited diversity, restricted to just a few option letters. Therefore, we replace it with a new simple prompt: \texttt{"Answer with the option's letter from the given choices first, and only after that, provide a detailed explanation for the choice."}. 

For each sample, the model generates multiple responses using random decoding. The correctness of each response is verified by matching it with the ground-truth answer—either the correct option letter or the "yes"/"no" word. A reward of $2$ is assigned to correct responses, while incorrect responses receive a reward of $0$. 
Furthermore, to ensure that the model follows the instructions and provides diverse responses for both multiple-choice and yes/no data, we add an extra format reward to the online data: if a response contains fewer than $\tau$ words, we apply a $\gamma$ penalty to the reward:

\begin{equation}
r_i
= 2 \cdot \1_\mathrm{\hat{y}_i = y_i}
  - \gamma \cdot \1_\mathrm{L_i < \tau} 
\end{equation}

where $\mathrm{\hat{y}_i}$ denotes the correct letter, $\mathrm{y_i}$ denotes the generated letter, and $\mathrm{L_i}$ denotes the number of words in the response. $\gamma$ and $\tau$ are two hyperparameters. In this way, we encourage the model to provide an explanation after the verifiable option letter, rather than generating only a single option letter.

Overall, to exploit the complementary strengths of both online and offline preference data, \MBPO{} integrates them into a unified hybrid preference dataset. During training, \MBPO{} randomly samples mini-batches from this combined dataset. For samples coming from the offline dataset, rewards are directly assigned to the chosen and rejected responses based on the known preference. For samples drawn from the online dataset, \MBPO{} first generates multiple candidate responses using the current policy model $\pi_\theta$, and then assigns rewards according to their agreement with the ground truth answer.

\section{Experiments}
\label{experiments}

In this section, we first introduce the implementation details, including training details, datasets, evaluation protocol and baseline methods. Subsequently, we present our main results comparing \MBPO{} with baseline methods on several general vision language tasks and hallucination benchmarks, demonstrating the effectiveness of \MBPO{}. In addition, the ablation study provides a closer look at \MBPO{} and verifies the contributions of its individual components. Lastly, we include additional experimental results for further analysis.

\subsection{Implementation Details}
\label{sec:implementation}

\textbf{Training details:} Following recent studies~\citep{shen2025vlm,chen2025r1v,zheng2025easyr1} that apply GRPO to train LMMs, we adopt \texttt{Qwen2-VL-7B-Instruct}~\citep{wang2024qwen2} and \texttt{Qwen2.5-VL-7B-Instruct}~\citep{Qwen2.5-VL} as our backbone models. The learning rate is set to $5 \times 10^{-7}$, and the KL-divergence coefficient ($\beta$) is set to $0.1$. Gradient accumulation is used to maintain an effective batch size of $16$. For each multiple-choice and yes/no sample, we generate $16$ responses to compute the GRPO advantage. A reward of $2$ is assigned to correct responses, and $0$ otherwise. $\gamma$ and $\tau$ are set to $0.5$ and $5$ respectively. For offline data, chosen responses are assigned a reward of $2$, while rejected responses receive reward $0$. To enable efficient training, we use \texttt{bfloat16} precision. For the adversarial image generation, we attack each image $20$ iterations and the step size $\alpha$ is set as $\frac{4}{255}$. All experiments are conducted using \texttt{PyTorch} and the Hugging Face \texttt{Transformers} library on $4\times$ NVIDIA H100 80GB GPUs.
             
\noindent\textbf{Datasets:} Following previous works~\citep{pi2024strengthening,zhou2024aligning}, we use high-quality visual instruction tuning data as our offline positive samples to train the powerful and up-to-date Qwen series models. Specifically, from the high-quality \texttt{MMSeed-163K} dataset~\citep{luo2024mmevol}, we randomly select $10K$ samples with high IIG for the offline dataset, along with all multiple-choice and yes/no samples (approximately 2K) as the online dataset. The \texttt{MMSeed-163K} dataset is a diverse multi-domain instruction dataset curated from LLaVA-Instruct~\citep{liu2023visual}, ShareGPT4V~\citep{chen2024sharegpt4v}, and Cambrian-1~\citep{tong2024cambrian}, encompassing 163K samples across tasks such as VQA, OCR, chart understanding and reasoning. More details can be found in the Appendix.

\noindent\textbf{Evaluation protocol:} We conduct a wide range of benchmarks to evaluate the comprehensive capabilities of LMMs, covering both general vision language tasks and hallucination benchmarks. For general vision language tasks, we use AI2D~\citep{kembhavi2016diagram}, MME~\citep{fu2023mme}, MMStar~\citep{chen2024we}, MMVet~\citep{yu2024mm} and MMBench~\citep{liu2024mmbench}. For hallucination benchmarks, we use MMHal-Bench~\citep{sun2023aligning} and ObjectHal~\citep{rohrbach2018object}. The evaluation is performed using the popular \texttt{LMMs-Eval} framework~\citep{zhang2024lmmsevalrealitycheckevaluation}. More details about these benchmarks can be found in the Appendix.

\noindent\textbf{Baselines:} We select studies that use preference learning to align LMMs as our baselines, including BPO~\citep{pi2024strengthening}, POVID~\citep{zhou2024aligning}, RLAIFV~\citep{yu2024rlaifv}, SIMA~\citep{wang2024enhancing}, CSR~\citep{zhou2024calibrated}, mDPO~\citep{wang2024mdpo}, MFPO~\citep{jiang2024modality}, FiSAO~\citep{cui2024fine}, and DAMA~\citep{lu2025damo}. For BPO, POVID, RLAIF-V, and CSR, we download their publicly released model weights and report evaluation results with the \texttt{LMMs-Eval} framework. For other methods, we report the results of the 7B model reported in their original papers. To ensure a fair comparison, we also train \texttt{Qwen2/2.5-VL-7B-Instruct} on the corresponding public datasets from BPO, POVID, RLAIF-V and CSR as additional baselines. More details about the baselines are provided in the Appendix.


\subsection{Benchmark Comparisons}
\label{main_exp}

In this section, we compare the performance of baseline methods and \MBPO{} on general vision-language tasks and hallucination benchmarks. The detailed results are presented in Table~\ref{tab:main_results}. If a baseline model is not available or the original paper does not report results on a specific benchmark, we use a “–” in the table.
On general vision-language tasks such as MME$\rm^p$, MMStar, and MMVet, \MBPO{} consistently outperforms all baselines with both Qwen base models. For example, \MBPO{} using \texttt{Qwen2-VL-7B} surpasses the second-best result on MME$\rm^p$ by $5.7$ points and on MMVet by $1.9$ points. When using the \texttt{Qwen2.5-VL-7B} backbone, \MBPO{} improves MMStar performance from $62.0$ to $63.0$, and MMVet from $62.2$ to $65.8$.
On the AI2D benchmark, which evaluates the factual knowledge of LMMs, all methods, including \MBPO{}, perform similarly and do not show significant improvements. This suggests that preference learning strategies cannot effectively enhance the factual knowledge of LMMs.
On hallucination benchmarks MMHal-Bench and ObjectHal, \MBPO{} achieves the best performance across most metrics. With \texttt{Qwen2-VL-7B}, \MBPO{} reduces CHAIR$\rm_{S}$ and CHAIR$\rm_{I}$ by $3.3$ and $1.6$ points respectively, compared to the base model. These reductions are even more pronounced with \texttt{Qwen2.5-VL-7B}, where CHAIR$\rm_{S}$ drops from $14.1$ to $7.4$, and CHAIR$\rm_{I}$ from $6.9$ to $3.6$, nearly halving the hallucination error. In addition, \MBPO{} improves MMHal$\rm^{score}$ from $3.68$ to $3.75$ and reduces MMHal$\rm^{rate}$ from $0.42$ to $0.34$, indicating fewer hallucinations in model responses.
In summary, \MBPO{} yields consistent and superior performance across a wide range of benchmarks based on the average of scores. It not only improves results on general vision-language tasks, but also significantly alleviates hallucination. These results highlight the advantage of encouraging LMMs to rely more on input visual information rather than the prior biases of the LLM backbone.

\begin{table}[t!]
\centering
\caption{Comparison with baseline methods on general vision language and hallucination benchmarks. \textsuperscript{*} indicates results reported in the original papers, and \textdownarrow{} indicates that lower is better. The best performance is marked in \textbf{bold}.}
\setlength{\tabcolsep}{4pt}
\renewcommand{\arraystretch}{1.2}
\resizebox{\textwidth}{!}{%
\begin{tabular}{lccccccc:cccc}
    \toprule
    \textbf{Model}
      & {\scriptsize\textbf{AI2D}}
      & {\scriptsize\textbf{MME\textsuperscript{p}}}
      & {\scriptsize\textbf{MMStar}}
      & {\scriptsize\textbf{MMVet}}
      & {\scriptsize\textbf{MMB}}
      & {\scriptsize\textbf{MMHal\textsuperscript{score}}}
      & {\scriptsize\textbf{Avg}}
      & {\scriptsize\textbf{MMHal\textsuperscript{rate}\textdownarrow}}
      & {\scriptsize\textbf{CHAIR\textsubscript{S}\textdownarrow}}
      & {\scriptsize\textbf{CHAIR\textsubscript{I}\textdownarrow}}
      & {\scriptsize\textbf{Avg}\textdownarrow{}}
      \\
    \midrule[0.6pt]
    BPO
      & --       & --     & --    & 36.8\textsuperscript{*}   & --    & -- & --    & --    & 31.9\textsuperscript{*}   & 15.1\textsuperscript{*} & --\\
    POVID
      & 54.2  & 1438.7 & 35.6  & 31.9      & 64.3  & 2.1  & 1626.8 & 0.60  & 37.9     & 18.9 & 57.4\\
    RLAIFV
      & 52.3  & 1356.0 & --    & 24.0      & 62.7  & 2.9   & -- & 0.46  & 8.6      & 4.3 & 13.4 \\
    SIMA
      & --    & 1507.7\textsuperscript{*}  & --    & 31.6\textsuperscript{*}   & 64.9\textsuperscript{*} & 2.3\textsuperscript{*} & -- & --    & 40.9\textsuperscript{*}   & 10.4\textsuperscript{*} & --\\
    CSR
      & 54.9  & 1523.3 & 34.3  & 31.1      & 64.1  & 2.2  & 1709.9 & 0.6   & 12.2     & 8.3 & 21.1 \\
    mDPO
      & --       & --     & --    & --        & --    & 2.39\textsuperscript{*} & -- & 0.54\textsuperscript{*} & 35.7\textsuperscript{*}  & 9.8\textsuperscript{*} & 46.1 \\
    MFPO
      & --       & --     & --    & --        & --    & 2.89\textsuperscript{*} & -- & 0.45\textsuperscript{*} & 10.6\textsuperscript{*}  & 5.1\textsuperscript{*} & 16.2 \\
    FiSAO
      & --   & 1522.6\textsuperscript{*} & --    & 30.7\textsuperscript{*} & 64.8\textsuperscript{*} & --    & -- & --    & 39.9\textsuperscript{*} & 9.9\textsuperscript{*} & -- \\
    DMMA
      & --       & --     & --    & 32.8\textsuperscript{*}   & --    & 2.76\textsuperscript{*} & -- & 0.41\textsuperscript{*} & --   & --    & -- \\
    \midrule[0.6pt]
    Qwen2-VL-7B
      & 80.4  & 1692.7 & 57.1  & 57.9      & 78.9  & 3.50  & 1970.5 & 0.34  & 10.9     & 5.9 & 17.1 \\
    {}+BPO
      & \textbf{80.6} & 1684.3 & 57.0  & 58.4      & 79.2  & 3.55  & 1963.1 & 0.31  & 8.7      & 4.8 & 13.8 \\
    {}+POVID
      & \textbf{80.6} & 1690.2 & \textbf{57.6}  & 58.9      & 78.6  & 3.53  & 1969.4 & \textbf{0.29} & 11.6     & 7.2 & 19.1\\
    {}+RLAIF-V
      & 80.4  & 1696.4 & 57.1  & 56.9      & 78.1  & 3.38  & 1972.3 & 0.34  & 9.2      & 5.6 & 15.1 \\
    {}+CSR
      & \textbf{80.6} & 1697.1 & 57.1  & 57.0      & 78.5  & 3.38 & 1973.7 & 0.35  & 21.4     & 11.6 & 33.4 \\
    \rowcolor{lightgray}
    {}\textbf{+\MBPO{} (ours)}
      & \textbf{80.6} & \textbf{1702.8} & \textbf{57.6} & \textbf{60.8} & \textbf{79.4} & \textbf{3.58} & \textbf{1984.5} & 0.36 & \textbf{7.6}  & \textbf{4.3} & \textbf{12.3} \\
    \midrule[0.6pt]
    Qwen2.5-VL-7B
      & 82.6  & 1680.1 & 62.0  & 62.2      & 83.2  & 3.68  & 1973.8 & 0.42  & 14.1     & 6.9 & 21.4\\
    {}+BPO
      & \textbf{82.7} & 1659.8 & 62.9  & 63.7      & 83.5  & 3.51  & 1956.1 & 0.42  & 9.9      & 5.4 & 15.7 \\
    {}+POVID
      & 82.6  & 1669.1 & 62.6  & 63.8      & 83.5  & 3.73  & 1965.3 & 0.37  & 10.5     & 5.7 & 16.6\\
    {}+RLAIF-V
      & \textbf{82.7} & 1686.3 & 62.7  & 63.8      & \textbf{83.6} & 3.63  & 1982.7 & 0.41  & 11.8     & 6.4 & 18.6 \\
    {}+CSR
      & 82.6   & 1687.8 & 62.1  & 61.7      & \textbf{83.6} & 3.71  & 1981.5 & 0.41  & 18.3     & 11.0 & 29.7 \\
    \rowcolor{lightgray}
    {}\textbf{+\MBPO{} (ours)}
      & 82.5  & \textbf{1706.3} & \textbf{63.0} & \textbf{65.8} & \textbf{83.6} & \textbf{3.75} & \textbf{2005.0} & \textbf{0.34} & \textbf{7.4} & \textbf{3.6} & \textbf{11.3} \\
    \bottomrule
\end{tabular}%
}
\vspace{-10pt}
\label{tab:main_results}
\end{table}


\subsection{Ablation Study}
\label{sec:ablation}

We conduct an ablation study on two Qwen base models across both general vision language tasks and hallucination benchmarks, following the same implementation details described in Section~\ref{sec:implementation}. To evaluate the effectiveness of each component in \MBPO{}, we incrementally add each one to the framework and measure its impact on each benchmark. The results are shown in Table~\ref{tab:ablation}, where \emph{+offline rand.} denotes offline rejected responses constructed using random noise sampled from $\mathcal{N}(0, 1)$. \emph{+offline adv.} indicates using only our offline dataset for training, and \emph{+online} refers to training the model solely on our online dataset. Based on the results, \MBPO{} achieves the best performance on $7$ out of $10$ benchmarks with \texttt{Qwen2-VL-7B} and on $6$ out of $10$ benchmarks with \texttt{Qwen2.5-VL-7B}. Furthermore, \MBPO{} performs the second best on $2$ of $10$ benchmarks with \texttt{Qwen2.5-VL-7B}. Thus, we conclude that each component of \MBPO{} is essential for achieving the best performance in most cases.

\begin{table}[t!]
\centering
\caption{Ablation studies of adding each component of \MBPO{} and their results on general vision language and hallucination benchmarks. \emph{+offline,rand.} indicates that the offline rejected samples are generated using images with random noise. We mark the best performance in \textbf{bold}.}
\setlength{\tabcolsep}{4pt}
\renewcommand{\arraystretch}{1.2}
\resizebox{\textwidth}{!}{%
\begin{tabular}{lccccccc:cccc}
    \toprule
    \textbf{Model}
      & {\scriptsize\textbf{AI2D}}
      & {\scriptsize\textbf{MME\textsuperscript{p}}}
      & {\scriptsize\textbf{MMStar}}
      & {\scriptsize\textbf{MMVet}}
      & {\scriptsize\textbf{MMB}}
      & {\scriptsize\textbf{MMHal\textsuperscript{score}}}
      & {\scriptsize\textbf{Avg}}
      & {\scriptsize\textbf{MMHal\textsuperscript{rate}\textdownarrow}}
      & {\scriptsize\textbf{CHAIR\textsubscript{S}\textdownarrow}}
      & {\scriptsize\textbf{CHAIR\textsubscript{I}\textdownarrow}}
      & {\scriptsize\textbf{Avg}\textdownarrow{}}\\
    \midrule[0.6pt]
    Qwen2-VL-7B
      & 80.4  & 1692.7 & 57.1  & 57.9      & 78.9  & 3.50  & 1970.5 & 0.34  & 10.9     & 5.9 & 17.1 \\
    {}+offline, rand.
      & \textbf{80.6} & 1684.8 & 57.8 & 58.5 & 78.6 & 3.54 & 1963.8 & 0.36 & 10.2 & 5.3 & 15.9\\
    {}+offline, adv.
      & 80.5 & 1697.6 & \textbf{58.0} & 59.8 & 78.8 & 3.50 & 1978.2 & \textbf{0.33} & 7.8 & 4.8 & 12.9\\
    {}+online, $\gamma=0$
      & 80.5  & 1682.3 & 57.5 & 59.0 & 78.4 & 3.46 & 1961.1 & 0.36 & 8.4 & 4.5 & 13.3\\
    {}+online, $\gamma=0.5$
      & 80.5  & 1681.9 & 57.4 & 60.6 & 78.4 & 3.52 & 1962.3 & 0.35 & 8.3 & \textbf{4.3} & 12.9\\
    \rowcolor{lightgray}
    {}\textbf{+\MBPO{}}
      & \textbf{80.6} & \textbf{1702.8} & 57.6 & \textbf{60.8} & \textbf{79.4} & \textbf{3.58} & \textbf{1984.5} & 0.36 & \textbf{7.6}  & \textbf{4.3} & \textbf{12.3}\\
    \midrule[0.6pt]
    Qwen2.5-VL-7B
      & 82.6  & 1680.1 & 62.0  & 62.2      & 83.2  & 3.68  & 1973.8 & 0.42  & 14.1     & 6.9 & 21.4\\
    {}+offline, rand.
      & 82.6 & 1688.4 & 62.1 & 61.5 & 83.4 & 3.57 & 1981.6 & 0.38 & 12.4 & 6.0 & 18.8 \\
    {}+offline, adv.
      & 82.5 & 1693.5 & 62.1 & 61.3 & 83.3 & 3.56 & 1986.3 & 0.38 & 7.5 & 4.1 & 12.0 \\
    {}+online, $\gamma=0$
      & \textbf{82.8}  & 1693.8 & 62.8 & 64.3 & 83.5 & 3.83 & 1991.0 & 0.39 & 10.6 & 6.1 & 17.1\\
    {}+online, $\gamma=0.5$
      & \textbf{82.8} & 1693.6 & 62.9 & 64.7 & \textbf{83.6} & \textbf{3.91} & 1991.5 & 0.37 & 9.8 & 5.4 & 15.6 \\
    \rowcolor{lightgray}
    {}\textbf{+\MBPO{}}
      & 82.5  & \textbf{1706.3} & \textbf{63.0} & \textbf{65.8} & \textbf{83.6} & 3.75 & \textbf{2005.0} & \textbf{0.34} & \textbf{7.4} & \textbf{3.6} & \textbf{11.3}\\
    \bottomrule
\end{tabular}%
}
\vspace{-10pt}
\label{tab:ablation}
\end{table}


\subsection{Further Analysis}

\begin{figure}[t!]
  \centering
  \includegraphics[width=1\linewidth]{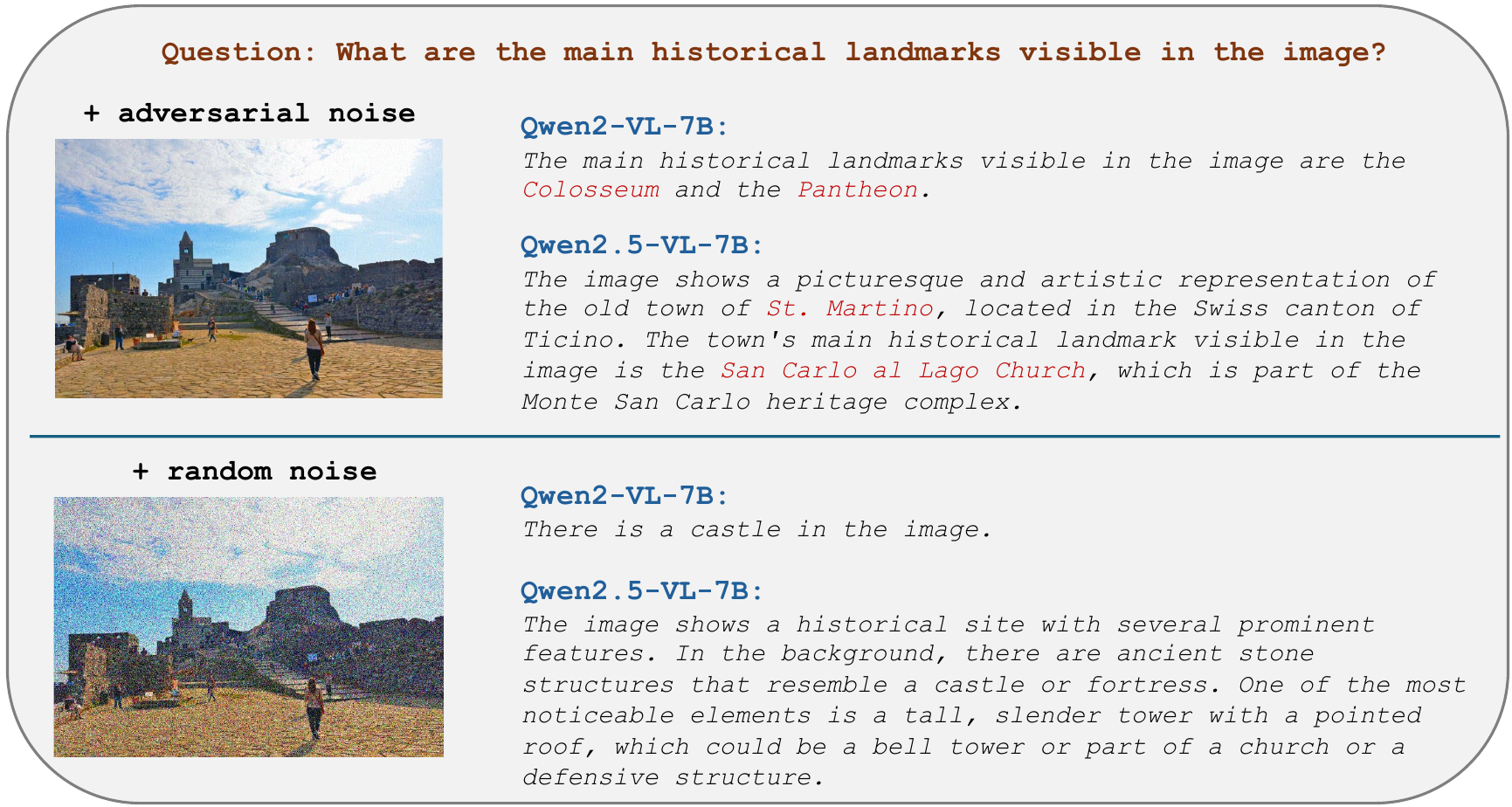} 
  \caption{An example comparing model responses of the image with adversarial noise and random noise. The prior bias from LLM is marked in \textcolor{red}{red}.}
  \label{fig:AdvImgExample}
\end{figure}

\textbf{Adversarial Image:} To gain a deeper understanding of the adversarial image, we present a detailed case study comparing model responses on an image with adversarial noise versus an image with random noise. The images and corresponding responses from Qwen2-VL-7B and Qwen2.5-VL-7B are shown in Fig.~\ref{fig:AdvImgExample}. For the adversarial image, both models follow the question's instruction and provide responses based on the prior biases of the LLM backbone. In contrast, for the image with random noise, both models fail to follow the question's instruction and instead offer a general description of the image, without leveraging the LLM's prior biases.

\begin{figure}[t]
    \centering
    \begin{minipage}[t]{0.48\textwidth}
        \centering
        \includegraphics[width=\linewidth]
        {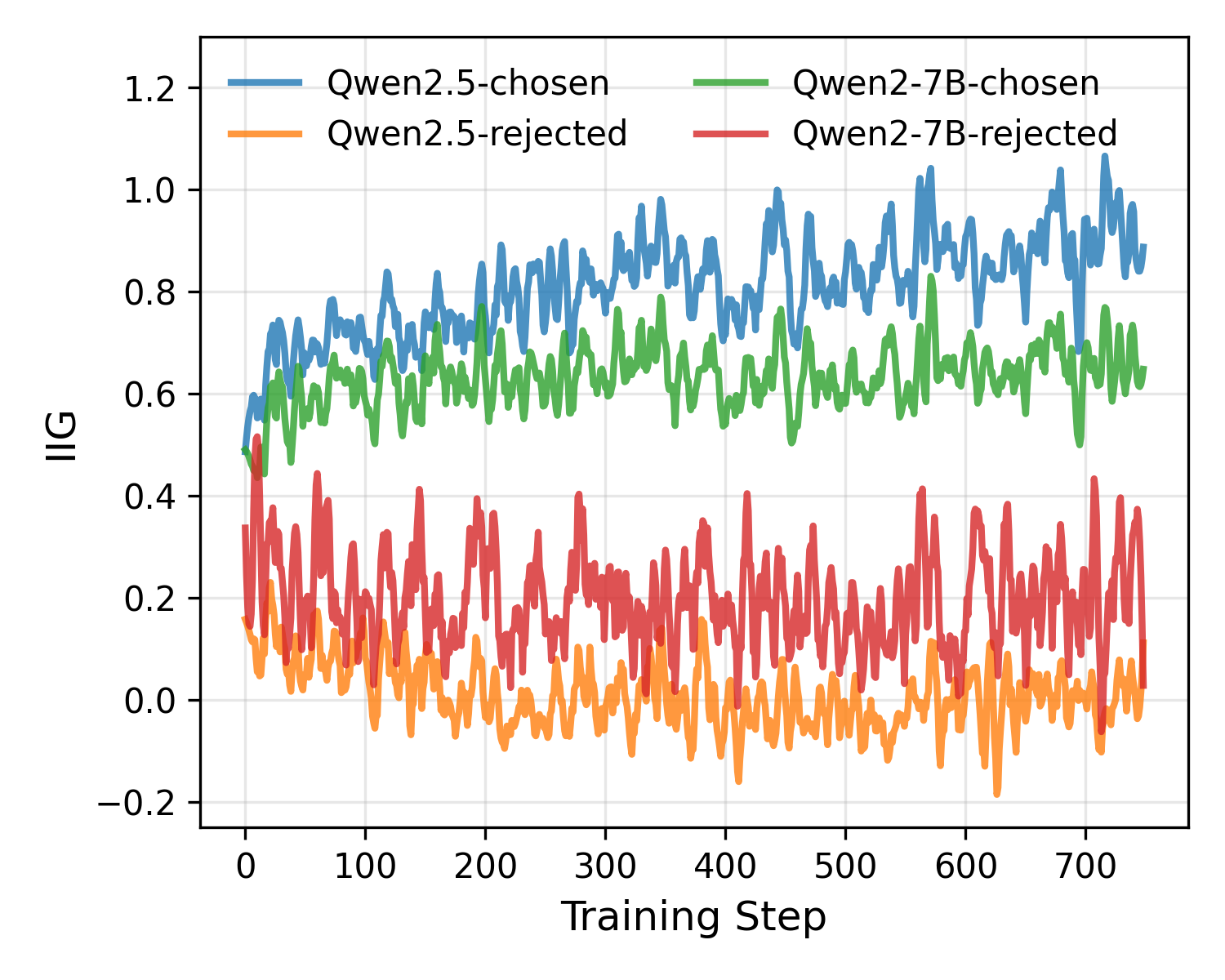}
        \caption{IIG of chosen and rejected responses change along with the training.}
        \label{fig:IIG}
    \end{minipage}
    \hfill
    \begin{minipage}[t]{0.48\textwidth}
        \centering
        \includegraphics[width=\linewidth]{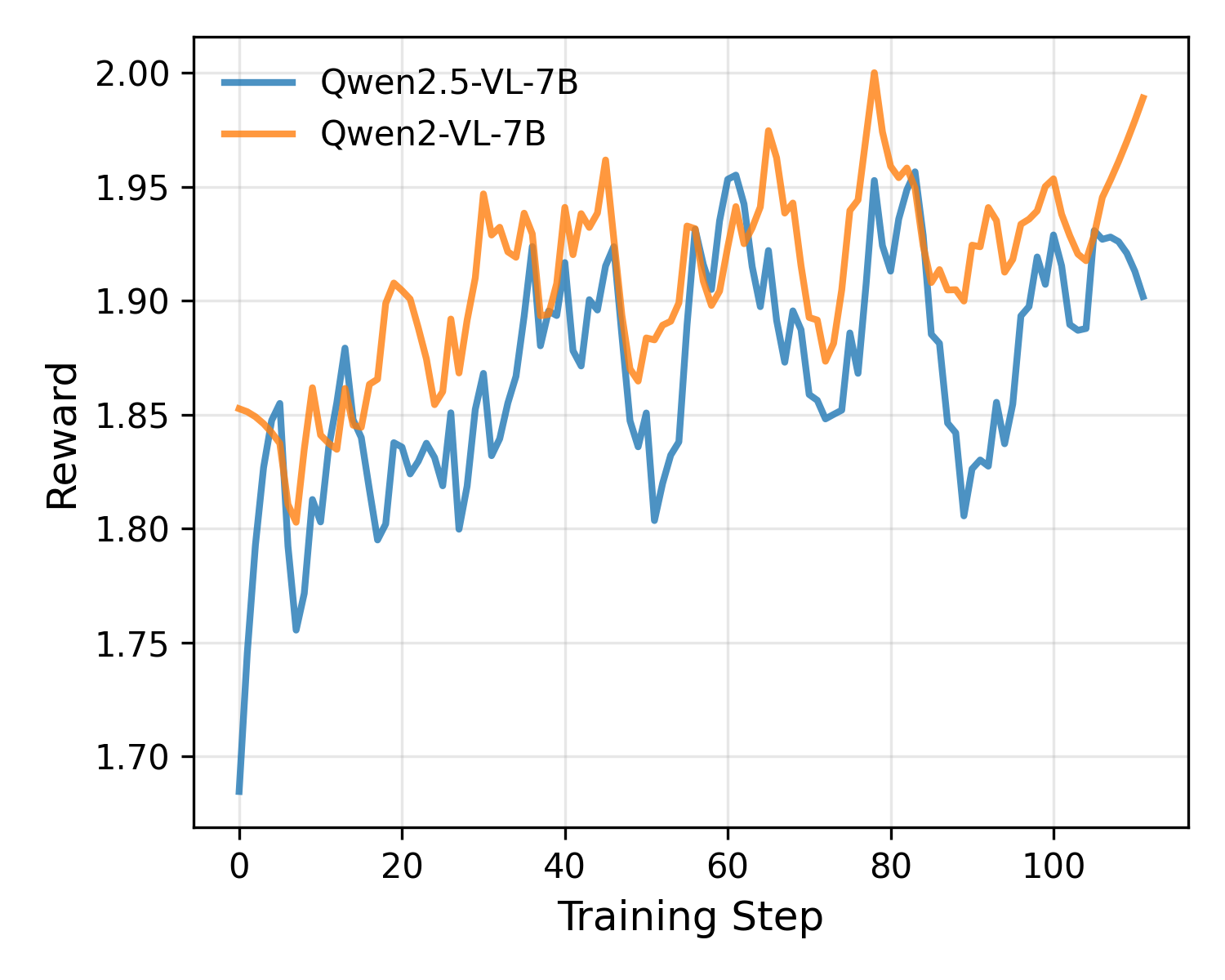}
        \caption{Reward of the online closed-end data changes along with the training.}
        \label{fig:reward}
    \end{minipage}
    \vspace{-2pt}
\end{figure}

\textbf{Image Information Gain:} The goal of \MBPO{} is to encourage LMMs to extract more information from the image, reflected by a higher IIG after training. Using the same \emph{+offline adv.} setting described in Section~\ref{sec:ablation}, we train the model on offline dataset and measure the change in IIG during training. The smoothed results are shown in Fig~\ref{fig:IIG}. As illustrated, the IIG of chosen responses increases throughout the training process, while the IIG of rejected responses remains consistently low. These results demonstrate that \MBPO{} effectively addresses the modality imbalance problem and successfully incorporates more visual information into the responses by training on our offline preference data.


\textbf{Closed-set Data Reward:}
To verify the effectiveness of learning from the online preference dataset, we measure the reward on closed-end data during training. The settings follow those of the \emph{+online} configuration in Section~\ref{sec:ablation}, where each model is trained on online closed-end data for one epoch. As shown in the smoothed results in Fig.~\ref{fig:reward}, the reward of closed-end data increases as training progresses for all models. This demonstrates the effectiveness of our online learning strategy, which improves model performance on closed-end questions through GRPO training.

\section{Conclusion}
\label{conclusion}

In this paper, we propose a new preference learning framework, Modality-Balancing Preference
Optimization (\MBPO{}) to address the modality imbalance problem. \MBPO{} optimizes the model with offline and online datasets in a hybrid manner. For the offline dataset, \MBPO{} mines rejected responses with limited visual information, thereby encouraging LMMs to incorporate more visual cues into their outputs. Additionally, we explore the potential to use closed-ended data as an online dataset and train with verified rewards using GRPO loss. Experimental results in both general vision language tasks and hallucination benchmarks demonstrate the effectiveness of \MBPO{} in aligning LMMs and addressing modality imbalance.

\clearpage  
\bibliographystyle{unsrt}  
\bibliography{references}  

\begin{thebibliography}{10}

\bibitem{liu2023visual}
Haotian Liu, Chunyuan Li, Qingyang Wu, and Yong~Jae Lee.
\newblock Visual instruction tuning.
\newblock {\em Advances in neural information processing systems}, 36:34892--34916, 2023.

\bibitem{liu2024llavanext}
Haotian Liu, Chunyuan Li, Yuheng Li, Bo~Li, Yuanhan Zhang, Sheng Shen, and Yong~Jae Lee.
\newblock Llava-next: Improved reasoning, ocr, and world knowledge, January 2024.

\bibitem{Qwen2.5-VL}
Shuai Bai, Keqin Chen, Xuejing Liu, Jialin Wang, Wenbin Ge, Sibo Song, Kai Dang, Peng Wang, Shijie Wang, Jun Tang, Humen Zhong, Yuanzhi Zhu, Mingkun Yang, Zhaohai Li, Jianqiang Wan, Pengfei Wang, Wei Ding, Zheren Fu, Yiheng Xu, Jiabo Ye, Xi~Zhang, Tianbao Xie, Zesen Cheng, Hang Zhang, Zhibo Yang, Haiyang Xu, and Junyang Lin.
\newblock Qwen2.5-vl technical report.
\newblock {\em arXiv preprint arXiv:2502.13923}, 2025.

\bibitem{tong2024cambrian}
Peter Tong, Ellis Brown, Penghao Wu, Sanghyun Woo, Adithya Jairam~Vedagiri IYER, Sai~Charitha Akula, Shusheng Yang, Jihan Yang, Manoj Middepogu, Ziteng Wang, et~al.
\newblock Cambrian-1: A fully open, vision-centric exploration of multimodal llms.
\newblock {\em Advances in Neural Information Processing Systems}, 37:87310--87356, 2024.

\bibitem{chen2024internvl}
Zhe Chen, Jiannan Wu, Wenhai Wang, Weijie Su, Guo Chen, Sen Xing, Muyan Zhong, Qinglong Zhang, Xizhou Zhu, Lewei Lu, et~al.
\newblock Internvl: Scaling up vision foundation models and aligning for generic visual-linguistic tasks.
\newblock In {\em Proceedings of the IEEE/CVF conference on computer vision and pattern recognition}, pages 24185--24198, 2024.

\bibitem{2023GPT4VisionSC}
Gpt-4v(ision) system card.
\newblock 2023.

\bibitem{xiong2024llavacritic}
Tianyi Xiong, Xiyao Wang, Dong Guo, Qinghao Ye, Haoqi Fan, Quanquan Gu, Heng Huang, and Chunyuan Li.
\newblock Llava-critic: Learning to evaluate multimodal models.
\newblock 2024.

\bibitem{chen2023gpt}
Ruibo Chen, Tianyi Xiong, Yihan Wu, Guodong Liu, Zhengmian Hu, Lichang Chen, Yanshuo Chen, Chenxi Liu, and Heng Huang.
\newblock Gpt-4 vision on medical image classification--a case study on covid-19 dataset.
\newblock {\em arXiv preprint arXiv:2310.18498}, 2023.

\bibitem{chen2024your}
Ruibo Chen, Yihan Wu, Lichang Chen, Guodong Liu, Qi~He, Tianyi Xiong, Chenxi Liu, Junfeng Guo, and Heng Huang.
\newblock Your vision-language model itself is a strong filter: Towards high-quality instruction tuning with data selection.
\newblock {\em arXiv preprint arXiv:2402.12501}, 2024.

\bibitem{liu2024few}
Chenxi Liu, Zhenyi Wang, Tianyi Xiong, Ruibo Chen, Yihan Wu, Junfeng Guo, and Heng Huang.
\newblock Few-shot class incremental learning with attention-aware self-adaptive prompt.
\newblock In {\em European Conference on Computer Vision}, pages 1--18. Springer, 2024.

\bibitem{li2024llava}
Bo~Li, Yuanhan Zhang, Dong Guo, Renrui Zhang, Feng Li, Hao Zhang, Kaichen Zhang, Peiyuan Zhang, Yanwei Li, Ziwei Liu, et~al.
\newblock Llava-onevision: Easy visual task transfer.
\newblock {\em arXiv preprint arXiv:2408.03326}, 2024.

\bibitem{luo2024mmevol}
Run Luo, Haonan Zhang, Longze Chen, Ting-En Lin, Xiong Liu, Yuchuan Wu, Min Yang, Minzheng Wang, Pengpeng Zeng, Lianli Gao, et~al.
\newblock Mmevol: Empowering multimodal large language models with evol-instruct.
\newblock {\em arXiv preprint arXiv:2409.05840}, 2024.

\bibitem{yu2024rlhf}
Tianyu Yu, Yuan Yao, Haoye Zhang, Taiwen He, Yifeng Han, Ganqu Cui, Jinyi Hu, Zhiyuan Liu, Hai-Tao Zheng, Maosong Sun, et~al.
\newblock Rlhf-v: Towards trustworthy mllms via behavior alignment from fine-grained correctional human feedback.
\newblock In {\em Proceedings of the IEEE/CVF Conference on Computer Vision and Pattern Recognition}, pages 13807--13816, 2024.

\bibitem{zhao2023beyond}
Zhiyuan Zhao, Bin Wang, Linke Ouyang, Xiaoyi Dong, Jiaqi Wang, and Conghui He.
\newblock Beyond hallucinations: Enhancing lvlms through hallucination-aware direct preference optimization.
\newblock {\em arXiv preprint arXiv:2311.16839}, 2023.

\bibitem{liu2024paying}
Shi Liu, Kecheng Zheng, and Wei Chen.
\newblock Paying more attention to image: A training-free method for alleviating hallucination in lvlms.
\newblock In {\em European Conference on Computer Vision}, pages 125--140. Springer, 2024.

\bibitem{jiang2024modality}
Songtao Jiang, Yan Zhang, Ruizhe Chen, Yeying Jin, and Zuozhu Liu.
\newblock Modality-fair preference optimization for trustworthy mllm alignment.
\newblock {\em arXiv preprint arXiv:2410.15334}, 2024.

\bibitem{zhou2024calibrated}
Yiyang Zhou, Zhiyuan Fan, Dongjie Cheng, Sihan Yang, Zhaorun Chen, Chenhang Cui, Xiyao Wang, Yun Li, Linjun Zhang, and Huaxiu Yao.
\newblock Calibrated self-rewarding vision language models.
\newblock {\em arXiv preprint arXiv:2405.14622}, 2024.

\bibitem{lu2025damo}
Jinda Lu, Junkang Wu, Jinghan Li, Xiaojun Jia, Shuo Wang, YiFan Zhang, Junfeng Fang, Xiang Wang, and Xiangnan He.
\newblock Damo: Data-and model-aware alignment of multi-modal llms.
\newblock {\em arXiv preprint arXiv:2502.01943}, 2025.

\bibitem{rafailov2023direct}
Rafael Rafailov, Archit Sharma, Eric Mitchell, Christopher~D Manning, Stefano Ermon, and Chelsea Finn.
\newblock Direct preference optimization: Your language model is secretly a reward model.
\newblock {\em Advances in Neural Information Processing Systems}, 36:53728--53741, 2023.

\bibitem{pi2024strengthening}
Renjie Pi, Tianyang Han, Wei Xiong, Jipeng Zhang, Runtao Liu, Rui Pan, and Tong Zhang.
\newblock Strengthening multimodal large language model with bootstrapped preference optimization.
\newblock In {\em European Conference on Computer Vision}, pages 382--398. Springer, 2024.

\bibitem{cui2024fine}
Chenhang Cui, An~Zhang, Yiyang Zhou, Zhaorun Chen, Gelei Deng, Huaxiu Yao, and Tat-Seng Chua.
\newblock Fine-grained verifiers: Preference modeling as next-token prediction in vision-language alignment.
\newblock {\em arXiv preprint arXiv:2410.14148}, 2024.

\bibitem{yu2024rlaifv}
Tianyu Yu, Haoye Zhang, Qiming Li, Qixin Xu, Yuan Yao, Da~Chen, Xiaoman Lu, Ganqu Cui, Yunkai Dang, Taiwen He, Xiaocheng Feng, Jun Song, Bo~Zheng, Zhiyuan Liu, Tat-Seng Chua, and Maosong Sun.
\newblock Rlaif-v: Open-source ai feedback leads to super gpt-4v trustworthiness.
\newblock {\em arXiv preprint arXiv:2405.17220}, 2024.

\bibitem{amirloo2024understanding}
Elmira Amirloo, Jean-Philippe Fauconnier, Christoph Roesmann, Christian Kerl, Rinu Boney, Yusu Qian, Zirui Wang, Afshin Dehghan, Yinfei Yang, Zhe Gan, et~al.
\newblock Understanding alignment in multimodal llms: A comprehensive study.
\newblock {\em arXiv preprint arXiv:2407.02477}, 2024.

\bibitem{chen2024self}
Zixiang Chen, Yihe Deng, Huizhuo Yuan, Kaixuan Ji, and Quanquan Gu.
\newblock Self-play fine-tuning converts weak language models to strong language models.
\newblock {\em arXiv preprint arXiv:2401.01335}, 2024.

\bibitem{chen2024optune}
Lichang Chen, Jiuhai Chen, Chenxi Liu, John Kirchenbauer, Davit Soselia, Chen Zhu, Tom Goldstein, Tianyi Zhou, and Heng Huang.
\newblock Optune: Efficient online preference tuning.
\newblock {\em arXiv preprint arXiv:2406.07657}, 2024.

\bibitem{shao2024deepseekmath}
Zhihong Shao, Peiyi Wang, Qihao Zhu, Runxin Xu, Junxiao Song, Xiao Bi, Haowei Zhang, Mingchuan Zhang, YK~Li, Y~Wu, et~al.
\newblock Deepseekmath: Pushing the limits of mathematical reasoning in open language models.
\newblock {\em arXiv preprint arXiv:2402.03300}, 2024.

\bibitem{guo2025deepseek}
Daya Guo, Dejian Yang, Haowei Zhang, Junxiao Song, Ruoyu Zhang, Runxin Xu, Qihao Zhu, Shirong Ma, Peiyi Wang, Xiao Bi, et~al.
\newblock Deepseek-r1: Incentivizing reasoning capability in llms via reinforcement learning.
\newblock {\em arXiv preprint arXiv:2501.12948}, 2025.

\bibitem{chen2025r1v}
Liang Chen, Lei Li, Haozhe Zhao, Yifan Song, and Vinci.
\newblock R1-v: Reinforcing super generalization ability in vision-language models with less than \$3.
\newblock \url{https://github.com/Deep-Agent/R1-V}, 2025.
\newblock Accessed: 2025-02-02.

\bibitem{shen2025vlm}
Haozhan Shen, Peng Liu, Jingcheng Li, Chunxin Fang, Yibo Ma, Jiajia Liao, Qiaoli Shen, Zilun Zhang, Kangjia Zhao, Qianqian Zhang, Ruochen Xu, and Tiancheng Zhao.
\newblock Vlm-r1: A stable and generalizable r1-style large vision-language model.
\newblock {\em arXiv preprint arXiv:2504.07615}, 2025.

\bibitem{zheng2025easyr1}
Yaowei Zheng, Junting Lu, Shenzhi Wang, Zhangchi Feng, Dongdong Kuang, and Yuwen Xiong.
\newblock Easyr1: An efficient, scalable, multi-modality rl training framework.
\newblock \url{https://github.com/hiyouga/EasyR1}, 2025.

\bibitem{ouyang2022training}
Long Ouyang, Jeffrey Wu, Xu~Jiang, Diogo Almeida, Carroll Wainwright, Pamela Mishkin, Chong Zhang, Sandhini Agarwal, Katarina Slama, Alex Ray, et~al.
\newblock Training language models to follow instructions with human feedback.
\newblock {\em Advances in neural information processing systems}, 35:27730--27744, 2022.

\bibitem{mcaleese2024llm}
Nat McAleese, Rai~Michael Pokorny, Juan Felipe~Ceron Uribe, Evgenia Nitishinskaya, Maja Trebacz, and Jan Leike.
\newblock Llm critics help catch llm bugs.
\newblock {\em arXiv preprint arXiv:2407.00215}, 2024.

\bibitem{sun2023aligning}
Zhiqing Sun, Sheng Shen, Shengcao Cao, Haotian Liu, Chunyuan Li, Yikang Shen, Chuang Gan, Liang-Yan Gui, Yu-Xiong Wang, Yiming Yang, et~al.
\newblock Aligning large multimodal models with factually augmented rlhf.
\newblock {\em arXiv preprint arXiv:2309.14525}, 2023.

\bibitem{li2023silkie}
Lei Li, Zhihui Xie, Mukai Li, Shunian Chen, Peiyi Wang, Liang Chen, Yazheng Yang, Benyou Wang, and Lingpeng Kong.
\newblock Silkie: Preference distillation for large visual language models.
\newblock {\em arXiv preprint arXiv:2312.10665}, 2023.

\bibitem{yuan2024selfrewarding}
Weizhe Yuan, Richard~Yuanzhe Pang, Kyunghyun Cho, Sainbayar Sukhbaatar, Jing Xu, and Jason Weston.
\newblock Self-rewarding language models, 2024.

\bibitem{wang2024enhancing}
Xiyao Wang, Jiuhai Chen, Zhaoyang Wang, Yuhang Zhou, Yiyang Zhou, Huaxiu Yao, Tianyi Zhou, Tom Goldstein, Parminder Bhatia, Furong Huang, et~al.
\newblock Enhancing visual-language modality alignment in large vision language models via self-improvement.
\newblock {\em arXiv preprint arXiv:2405.15973}, 2024.

\bibitem{skalse2022defining}
Joar Skalse, Nikolaus Howe, Dmitrii Krasheninnikov, and David Krueger.
\newblock Defining and characterizing reward gaming.
\newblock {\em Advances in Neural Information Processing Systems}, 35:9460--9471, 2022.

\bibitem{zhou2024aligning}
Yiyang Zhou, Chenhang Cui, Rafael Rafailov, Chelsea Finn, and Huaxiu Yao.
\newblock Aligning modalities in vision large language models via preference fine-tuning.
\newblock {\em arXiv preprint arXiv:2402.11411}, 2024.

\bibitem{wang2024mdpo}
Fei Wang, Wenxuan Zhou, James~Y Huang, Nan Xu, Sheng Zhang, Hoifung Poon, and Muhao Chen.
\newblock mdpo: Conditional preference optimization for multimodal large language models.
\newblock {\em arXiv preprint arXiv:2406.11839}, 2024.

\bibitem{liu2025noisyrollout}
Xiangyan Liu, Jinjie Ni, Zijian Wu, Chao Du, Longxu Dou, Haonan Wang, Tianyu Pang, and Michael~Qizhe Shieh.
\newblock Noisyrollout: Reinforcing visual reasoning with data augmentation.
\newblock {\em arXiv preprint arXiv:2504.13055}, 2025.

\bibitem{jaech2024openai}
Aaron Jaech, Adam Kalai, Adam Lerer, Adam Richardson, Ahmed El-Kishky, Aiden Low, Alec Helyar, Aleksander Madry, Alex Beutel, Alex Carney, et~al.
\newblock Openai o1 system card.
\newblock {\em arXiv preprint arXiv:2412.16720}, 2024.

\bibitem{team2025kimi}
Kimi Team, Angang Du, Bofei Gao, Bowei Xing, Changjiu Jiang, Cheng Chen, Cheng Li, Chenjun Xiao, Chenzhuang Du, Chonghua Liao, et~al.
\newblock Kimi k1. 5: Scaling reinforcement learning with llms.
\newblock {\em arXiv preprint arXiv:2501.12599}, 2025.

\bibitem{meng2025mm}
Fanqing Meng, Lingxiao Du, Zongkai Liu, Zhixiang Zhou, Quanfeng Lu, Daocheng Fu, Tiancheng Han, Botian Shi, Wenhai Wang, Junjun He, et~al.
\newblock Mm-eureka: Exploring the frontiers of multimodal reasoning with rule-based reinforcement learning.
\newblock {\em arXiv preprint arXiv:2503.07365}, 2025.

\bibitem{huang2025vision}
Wenxuan Huang, Bohan Jia, Zijie Zhai, Shaosheng Cao, Zheyu Ye, Fei Zhao, Zhe Xu, Yao Hu, and Shaohui Lin.
\newblock Vision-r1: Incentivizing reasoning capability in multimodal large language models.
\newblock {\em arXiv preprint arXiv:2503.06749}, 2025.

\bibitem{peng2025lmm}
Yingzhe Peng, Gongrui Zhang, Miaosen Zhang, Zhiyuan You, Jie Liu, Qipeng Zhu, Kai Yang, Xingzhong Xu, Xin Geng, and Xu~Yang.
\newblock Lmm-r1: Empowering 3b lmms with strong reasoning abilities through two-stage rule-based rl.
\newblock {\em arXiv preprint arXiv:2503.07536}, 2025.

\bibitem{yang2025r1}
Yi~Yang, Xiaoxuan He, Hongkun Pan, Xiyan Jiang, Yan Deng, Xingtao Yang, Haoyu Lu, Dacheng Yin, Fengyun Rao, Minfeng Zhu, et~al.
\newblock R1-onevision: Advancing generalized multimodal reasoning through cross-modal formalization.
\newblock {\em arXiv preprint arXiv:2503.10615}, 2025.

\bibitem{yu2025perception}
En~Yu, Kangheng Lin, Liang Zhao, Jisheng Yin, Yuang Peng, Haoran Wei, Jianjian Sun, Chunrui Han, Zheng Ge, Xiangyu Zhang, Daxin Jiang, Jingyu Wang, and Wenbing Tao.
\newblock Perception r1: Pioneering perception policy with reinforcement learning.
\newblock {\em arXiv preprint arXiv:2504.07954}, 2025.

\bibitem{zhan2025vision}
Yufei Zhan, Yousong Zhu, Shurong Zheng, Hongyin Zhao, Fan Yang, Ming Tang, and Jinqiao Wang.
\newblock Vision-r1: Evolving human-free alignment in large vision-language models via vision-guided reinforcement learning.
\newblock {\em arXiv preprint arXiv:2503.18013}, 2025.

\bibitem{liu2025seg}
Yuqi Liu, Bohao Peng, Zhisheng Zhong, Zihao Yue, Fanbin Lu, Bei Yu, and Jiaya Jia.
\newblock Seg-zero: Reasoning-chain guided segmentation via cognitive reinforcement.
\newblock {\em arXiv preprint arXiv:2503.06520}, 2025.

\bibitem{madry2017towards}
Aleksander Madry, Aleksandar Makelov, Ludwig Schmidt, Dimitris Tsipras, and Adrian Vladu.
\newblock Towards deep learning models resistant to adversarial attacks.
\newblock {\em arXiv preprint arXiv:1706.06083}, 2017.

\bibitem{goodfellow2014explaining}
Ian~J Goodfellow, Jonathon Shlens, and Christian Szegedy.
\newblock Explaining and harnessing adversarial examples.
\newblock {\em arXiv preprint arXiv:1412.6572}, 2014.

\bibitem{wang2024qwen2}
Peng Wang, Shuai Bai, Sinan Tan, Shijie Wang, Zhihao Fan, Jinze Bai, Keqin Chen, Xuejing Liu, Jialin Wang, Wenbin Ge, et~al.
\newblock Qwen2-vl: Enhancing vision-language model's perception of the world at any resolution.
\newblock {\em arXiv preprint arXiv:2409.12191}, 2024.

\bibitem{chen2024sharegpt4v}
Lin Chen, Jinsong Li, Xiaoyi Dong, Pan Zhang, Conghui He, Jiaqi Wang, Feng Zhao, and Dahua Lin.
\newblock Sharegpt4v: Improving large multi-modal models with better captions.
\newblock In {\em European Conference on Computer Vision}, pages 370--387. Springer, 2024.

\bibitem{kembhavi2016diagram}
Aniruddha Kembhavi, Mike Salvato, Eric Kolve, Minjoon Seo, Hannaneh Hajishirzi, and Ali Farhadi.
\newblock A diagram is worth a dozen images.
\newblock In {\em Computer Vision--ECCV 2016: 14th European Conference, Amsterdam, The Netherlands, October 11--14, 2016, Proceedings, Part IV 14}, pages 235--251. Springer, 2016.

\bibitem{fu2023mme}
Chaoyou Fu, Peixian Chen, Yunhang Shen, Yulei Qin, Mengdan Zhang, Xu~Lin, Jinrui Yang, Xiawu Zheng, Ke~Li, Xing Sun, et~al.
\newblock Mme: A comprehensive evaluation benchmark for multimodal large language models.
\newblock {\em arXiv preprint arXiv:2306.13394}, 2023.

\bibitem{chen2024we}
Lin Chen, Jinsong Li, Xiaoyi Dong, Pan Zhang, Yuhang Zang, Zehui Chen, Haodong Duan, Jiaqi Wang, Yu~Qiao, Dahua Lin, et~al.
\newblock Are we on the right way for evaluating large vision-language models?
\newblock {\em arXiv preprint arXiv:2403.20330}, 2024.

\bibitem{yu2024mm}
Weihao Yu, Zhengyuan Yang, Linjie Li, Jianfeng Wang, Kevin Lin, Zicheng Liu, Xinchao Wang, and Lijuan Wang.
\newblock Mm-vet: Evaluating large multimodal models for integrated capabilities.
\newblock In {\em International conference on machine learning}. PMLR, 2024.

\bibitem{liu2024mmbench}
Yuan Liu, Haodong Duan, Yuanhan Zhang, Bo~Li, Songyang Zhang, Wangbo Zhao, Yike Yuan, Jiaqi Wang, Conghui He, Ziwei Liu, et~al.
\newblock Mmbench: Is your multi-modal model an all-around player?
\newblock In {\em European conference on computer vision}, pages 216--233. Springer, 2024.

\bibitem{rohrbach2018object}
Anna Rohrbach, Lisa~Anne Hendricks, Kaylee Burns, Trevor Darrell, and Kate Saenko.
\newblock Object hallucination in image captioning.
\newblock {\em arXiv preprint arXiv:1809.02156}, 2018.

\bibitem{zhang2024lmmsevalrealitycheckevaluation}
Kaichen Zhang, Bo~Li, Peiyuan Zhang, Fanyi Pu, Joshua~Adrian Cahyono, Kairui Hu, Shuai Liu, Yuanhan Zhang, Jingkang Yang, Chunyuan Li, and Ziwei Liu.
\newblock Lmms-eval: Reality check on the evaluation of large multimodal models, 2024.

\end{thebibliography}

\clearpage  
\appendix
\begin{center}
    {\LARGE \bf Appendix}
\end{center}

\section{Experiemntal Details}
\subsection{Dataset}
The \texttt{MMSeed-163K} dataset~\citep{luo2024mmevol} is a curated collection of $163K$ high-quality image-text instruction samples designed to support multimodal language model training. It integrates and refines data from \texttt{LLaVA-Instruct}~\citep{liu2023visual}, \texttt{ShareGPT4V}~\citep{chen2024sharegpt4v}, and \texttt{Cambrain-1}~\citep{tong2024cambrian}, covering diverse instruction formats including dialogue-based QA, global descriptions, scientific reasoning, and chart interpretation. As the goal of \MBPO{} is to balance modality in \emph{Large Multimodal Models} (LMMs) by encouraging more visual information to be used, we use responses with high IIG as the chosen responses in our offline dataset. To construct our offline dataset efficiently, we first randomly select $60K$ samples without closed-end questions from the \texttt{MMSeed-163K}. Then we use \texttt{Qwen2-VL-2B}~\citep{wang2024qwen2} to compute the \emph{Image Information Gain} (IIG) of each sample and choose $10K$ samples with the highest IIG as our offline dataset.

\subsection{Evaluation Benchmarks}
\begin{itemize}[leftmargin=2em]
    \setlength{\itemsep}{0.3em}
    \item \textbf{AI2D}~\citep{kembhavi2016diagram} is a large-scale dataset designed to evaluate a model’s ability to interpret and reason about grade school science diagrams. It contains over 5,000 annotated diagrams with more than 150,000 detailed annotations, syntactic parses, and 15,000+ multiple-choice questions. The benchmark focuses on two key tasks: Syntactic Parsing, which involves detecting diagram components and their structural relationships, and Semantic Interpretation, which maps these components to real-world concepts and events. 
    
    \item \textbf{MME}~\citep{fu2023mme} is a comprehensive benchmark designed to evaluate LMMs across two core dimensions: \emph{perception} (MME$\rm^p$) and \emph{cognition} (MME$\rm^c$). It consists of 14 subtasks, each crafted to assess a model's ability to interpret visual content and reason about it. For each image, the benchmark poses two questions whose answers are marked yes [Y] and no [N], respectively, allowing for a fine-grained evaluation of LMMs.

    \item \textbf{MMStar}~\citep{chen2024we} is a high-quality vision-indispensable benchmark designed to rigorously evaluate the multimodal capabilities of LMMs. It comprises 1,500 human-curated samples across 6 core capabilities and 18 fine-grained evaluation axes, offering a comprehensive and balanced assessment of models' understanding of both visual and textual modalities.

    \item \textbf{MMVet}~\citep{yu2024mm} is a comprehensive benchmark designed to evaluate the integration capabilities of generalist vision-language models. It defines six core VL abilities and systematically examines sixteen meaningful pairwise combinations to assess how well models can jointly reason over multiple modalities. To address the challenge of evaluating open-ended outputs, MMVet introduces an LLM-based evaluator. Specifically, we use the OpenAI API \emph{gpt-4o-2024-08-06} as our evaluator model.

    \item \textbf{MMBench}~\citep{liu2024mmbench} is a comprehensive benchmark designed to objectively and systematically evaluate the capabilities of LMMs. It consists of over 3,000 multiple-choice questions spanning 20 ability dimensions, including object localization, social reasoning, and more. Each dimension includes approximately 125 questions, ensuring balanced coverage across various vision-language skills.

    \item \textbf{MMHal-Bench}~\citep{sun2023aligning} is a benchmark designed to evaluate hallucinations in large multimodal models (LMMs) through 96 adversarially constructed image-question pairs. These pairs span 8 hallucination types and cover 12 object topics from COCO. A GPT model (OpenAI \emph{gpt-4o-2024-08-06}) is used as an evaluator by providing it with the image category, the question, the LMM’s response, and a human-generated reference answer. The overall score and hallucination rate are reported to measure the model performance on MMHal-Bench.
    
    \item \textbf{ObjectHal}~\citep{rohrbach2018object} is a widely adopted benchmark for assessing common object hallucination in detailed image descriptions. Following~\cite{yu2024rlaifv}, we employ 8 diverse prompts per image to improve evaluation stability. It assesses object
 hallucination at the instance and sentence levels, which can be calculated as:  
    \begin{equation}
    \text{CHAIR}_I = \frac{|\{\text{hallucinated objects}\}|}{|\{\text{all mentioned objects}\}|}
    \qquad
    \text{CHAIR}_S = \frac{|\{\text{captions with hallucinated objects}\}|}{|\{\text{all captions}\}|}
    \end{equation}
\end{itemize}

\subsection{Baselines}
\begin{itemize}[leftmargin=2em]
    \setlength{\itemsep}{0.3em}
    \item \textbf{BPO}~\citep{pi2024strengthening} generates negative responses directly from the model to perform preference learning. It introduces two key strategies: (1) using distorted images to trigger language-biased outputs, and (2) using a text-only LLM to inject common but incorrect elements into otherwise correct responses. These bootstrapped negatives are paired with high-quality references to train the model via preference optimization.
    
    \item \textbf{POVID}~\citep{zhou2024aligning} uses ground-truth instructions as preferred responses, and creates dispreferred responses through two different hallucination strategies: (1) prompting GPT-4V to inject plausible hallucinations into correct answers, and (2) distorting input images to elicit hallucinations from the VLM itself. These pairwise preference samples are then trained with Direct Preference Optimization (DPO).
    
    \item \textbf{RLAIF-V}~\citep{yu2024rlaifv} introduces two key innovations to enhance reward learning from AI feedback. First, it improves feedback quality by generating candidate responses through multiple decoding trials under identical conditions, effectively removing confounding factors like text style. It also uses a divide-and-conquer strategy to break complex response evaluation into simpler claim-level judgments, enabling more accurate and efficient preference modeling. Second, for inference-time guidance, RLAIF-V employs a self-feedback mechanism using reward scores from models aligned via Direct Preference Optimization (DPO) to refine responses without external supervision.
    
    \item \textbf{SIMA}~\citep{wang2024enhancing} leverages existing vision instruction datasets to self-generate responses and uses an in-context self-critic mechanism to create preference pairs for tuning. By designing specialized critic prompts, SIMA enables the LMM itself to act as the judge, eliminating the need for extra fine-tuning. Additionally, it introduces three new visual metrics to guide the self-critique process, boosting the reliability of preference judgments.
    
    \item \textbf{CSR}~\citep{zhou2024calibrated} enables the model to refine itself by repeatedly generating candidate responses, scoring each with a reward function, and compiling the highest-rated examples into preference data for fine-tuning. In its reward-modeling phase, CSR follows a step-wise strategy and embeds visual constraints within the self-rewarding process to amplify the impact of visual signals.
    
    \item \textbf{mDPO}~\citep{wang2024mdpo} aligns LMMs by optimizing image preference data, rather than relying solely on text-based preference. To stabilize training, MDPO introduces a reward anchor that ensures chosen responses always receive positive rewards, mitigating the risk of degrading their likelihood.
    
    \item \textbf{MFPO}~\citep{jiang2024modality} constructs image preference data by identifying hallucination-prone regions via keyword extraction and mapping them to image segments using the Segment Anything Model. Fine-grained noisy images are used as negative samples, and a reward function is built to favor clean over noisy regions. MFPO also incorporates a curriculum learning-inspired hierarchical alignment strategy that categorizes training data by difficulty (easy to hard), enabling stable and progressive learning. Margin loss is used to ensure consistent reward separation between preferred and rejected responses.
    
    \item \textbf{FiSAO}~\citep{cui2024fine} is a self-alignment approach for LMMs that enhances multimodal alignment without requiring extra data. It leverages the model’s own vision encoder as a fine-grained verifier to provide token-level feedback during training. This enables more precise supervision and improves alignment performance beyond traditional preference tuning methods.
    
    \item \textbf{DAMA}~\citep{lu2025damo} dynamically adjusts the preference optimization coefficient $\beta$ based on both data hardness and the model's responsiveness. It measures the difficulty based on CLIP-based image-text similarity. Furthermore, it adapts $\beta$ based on real-time responsiveness inferred from reward gaps between preferred and rejected responses. This dual adaptation allows DAMA to improve model alignment by preventing both overfitting on easy samples and underfitting on hard ones.

\end{itemize}

\section{Additional Experiments}
\subsection{Adversarial Noise}

We conduct additional experiments to explore the impact of iteration and step size in generating adversarial noise. As shown in Table~\ref{tab:adv}, we report experimental results using \texttt{Qwen2-VL-7B} as the base model, and compare the performance of different iteraion and step size pairs. In the table, \emph{+$(i,j)$} means we construct offline dataset by adversarial attack on the image $i$ iterations with $\frac{j}{255}$ as each step size. The training setting is the same as the \emph{+offline adv.} in our ablation study. Based on the results, we can see that the adversarial attack performs similarly on all benchmarks except for the MME and ObjectHal.

\begin{table}[t!]
\centering
\caption{Exploration of the impact of iteration and step size in generating adversarial images for the offline dataset. \emph{+$(i,j)$} stands for $i$ iterations and $\frac{j}{255}$ step size. We mark the best performance \textbf{bold}.}
\setlength{\tabcolsep}{4pt}
\renewcommand{\arraystretch}{1.2}
\resizebox{\textwidth}{!}{%
\begin{tabular}{lcccccccccc}
    \toprule
    \textbf{Model}
      & \textbf{\small{AI2D}}
      & \textbf{\small{MME$\rm^c$}}
      & \textbf{\small{MME$\rm^p$}}
      & \textbf{\small{MMStar}}
      & \textbf{\small{MMVet}}
      & \textbf{\small{MMB}}
      & \textbf{\small{MMHal$\rm^{score}$}}
      & \textbf{\small{MMHal$\rm^{rate}\downarrow$}}
      & \textbf{\small{CHAIR$\rm_{S}\downarrow$}}
      & \textbf{\small{CHAIR$\rm_{I}\downarrow$}} \\
    \midrule[0.6pt]
    Qwen2-VL-7B     & 80.4 & 628.2 & 1692.7 & 57.1 & 57.9  & 78.9 & 3.50 & 0.34 & 10.9 & 5.9 \\
    +(5,4)       & 80.5   & 635.7    & 1704.9     & 57.5   & \textbf{60.0}     & 78.9   & \textbf{3.57}   & 0.40    & \textbf{4.5} & \textbf{2.5} \\
    
    +(10,4)       & \textbf{80.7}   & 637.8    & 1701.7     &  57.9  &  57.5    &  79.0  & 3.54   & 0.40    & 6.4 & 3.5 \\

    +(20,2)       & 80.5   & \textbf{640.0}    & \textbf{1706.5}     & 57.7   & 59.4     & 78.8   & \textbf{3.57}   & 0.39    & 7.6 & 4.0 \\
    
    +(20,4) & 80.5 & 635.7 & 1697.6 & \textbf{58.0} & 59.8  & 78.8 & 3.50 & \textbf{0.33} & 7.8 & 4.8 \\

    +(20,8)       & \textbf{80.7}   & 628.2    & 1700.4     & 57.7   & 59.3     & \textbf{79.3}   & 3.39   & 0.41    & 7.0 & 3.9 \\
    \bottomrule
\end{tabular}%
}
\vspace{-10pt}
\label{tab:adv}
\end{table}


\subsection{Case Study}
\begin{figure}[t!]
  \centering
  \includegraphics[width=1\linewidth]{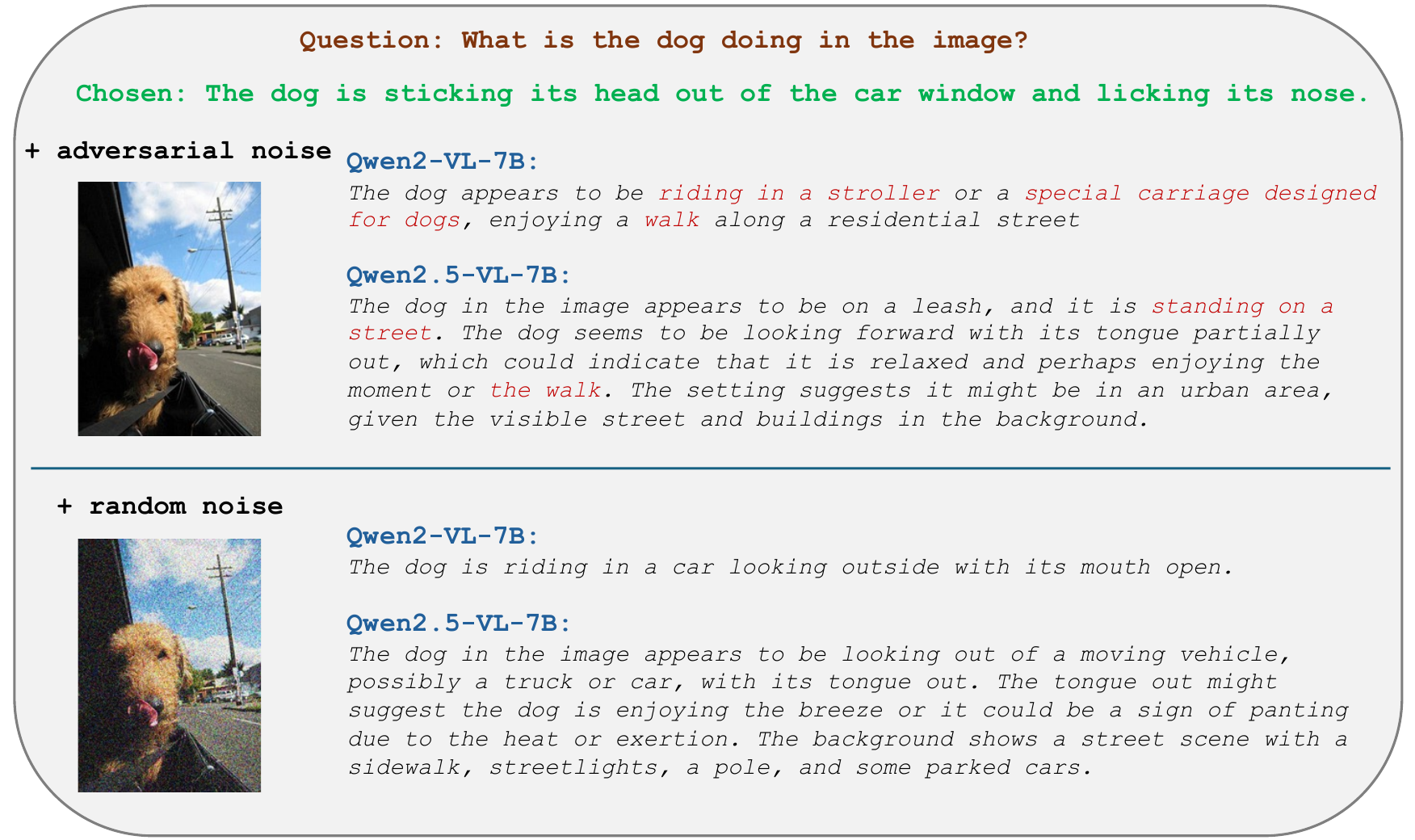} 
  \caption{An example comparing model responses of the image with adversarial noise and random noise. The prior bias from LLM is marked in \textcolor{red}{red}.}
  \label{fig:Appendix_adv}
\end{figure}

\begin{figure}[t!]
  \centering
  \includegraphics[width=0.8\linewidth]{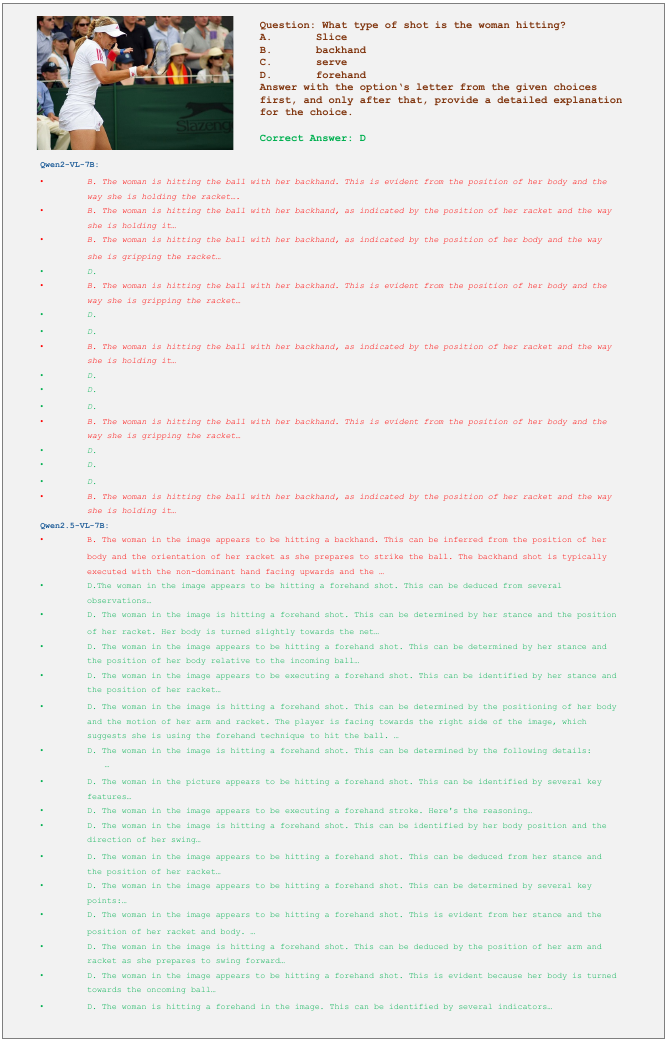} 
  \caption{An example of model responses on a multiple-choice question. Explanations are hidden for clarity. Each model generates $16$ responses using random decoding. The correct responses from LMM are marked in \textcolor{green}{green}.}
  \label{fig:Appendix_mulitple_choice}
\end{figure}

\begin{figure}[t!]
  \centering
  \includegraphics[width=0.8\linewidth]{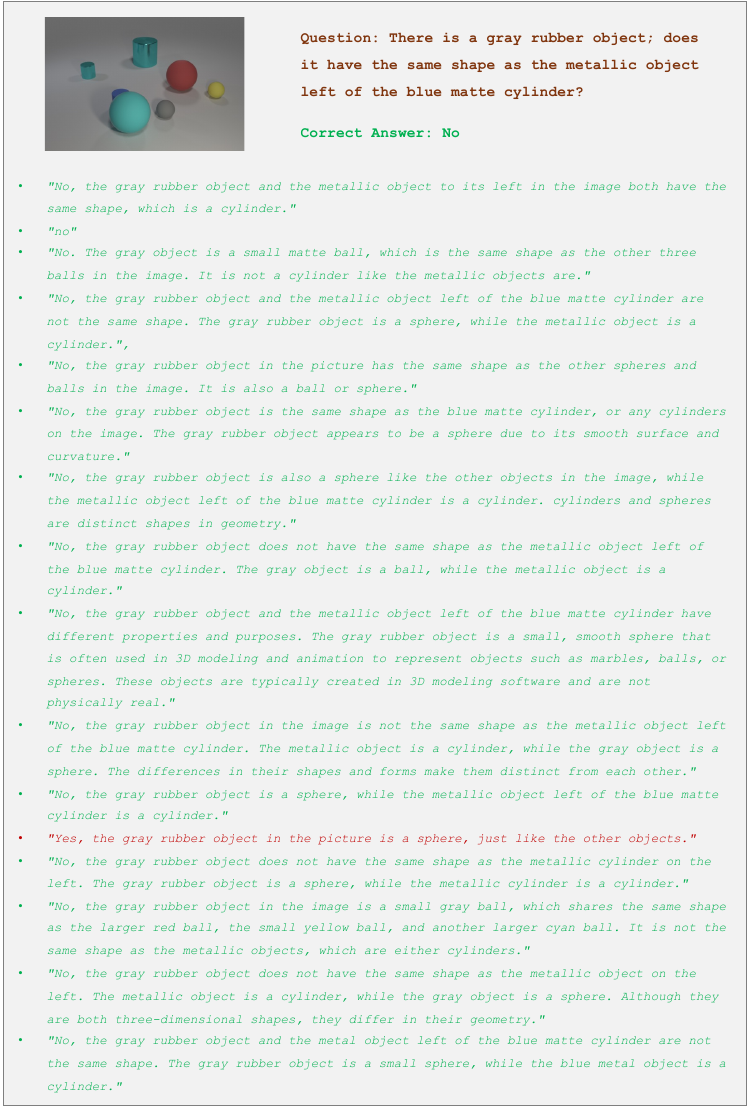} 
  \caption{An example of responses from \texttt{Qwen2-VL-7B} on a Yes/No question. The model generates $16$ responses using random decoding. The incorrect responses from the model are marked in \textcolor{red}{red}.}
  \label{fig:Appendix_YesNo_Qwen2}
\end{figure}

\begin{figure}[t!]
  \centering
  \includegraphics[width=0.9\linewidth]{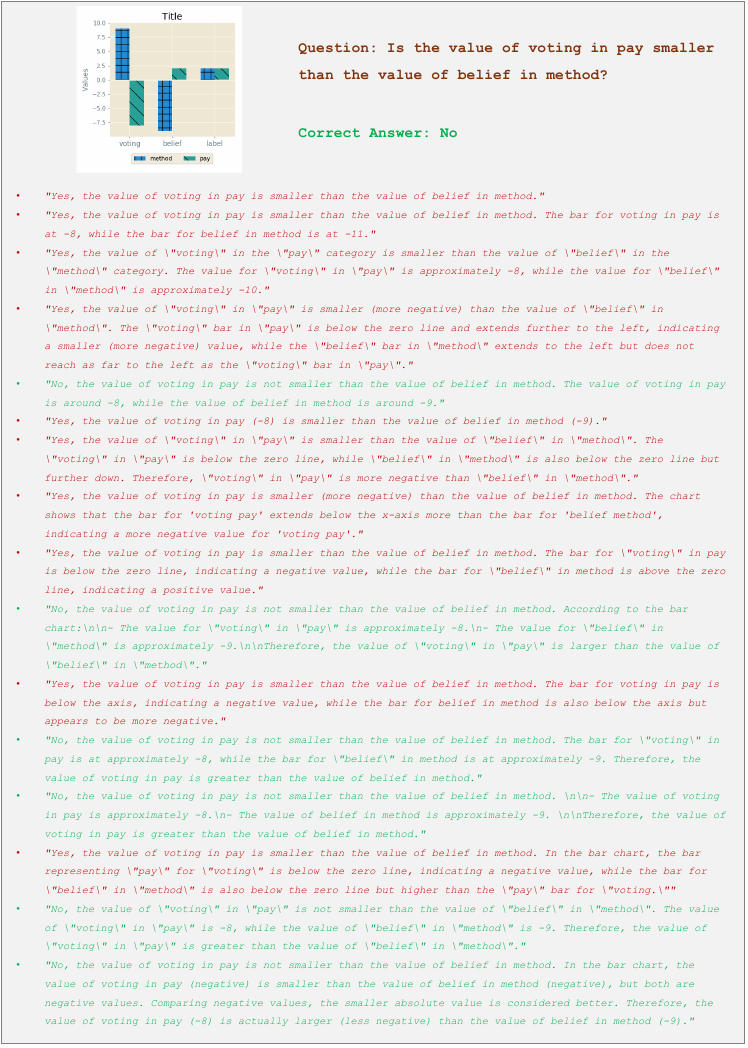} 
  \caption{An example of responses from \texttt{Qwen2.5-VL-7B} on a Yes/No question. The model generates $16$ responses using random decoding. The incorrect responses from the model are marked in \textcolor{red}{red}.}
  \label{fig:Appendix_YesNo_Qwen2_5}
\end{figure}

In this section, we provide some detailed case studies of the model output for both online and offline datasets. In Fig.~\ref{fig:Appendix_adv}, we show an example to compare model responses on images with adversarial noise and random noise. The image with adversarial noise effectively triggers the prior biases of the LLM backbone. Furthermore, an example of model's responses to an online multiple-choice question is shown in Fig.~\ref{fig:Appendix_mulitple_choice}. The response from \texttt{Qwen2-VL-7B} and \texttt{Qwen2.5-VL-7B} on online Yes/No data are shown in Fig.~\ref{fig:Appendix_YesNo_Qwen2} and Fig.~\ref{fig:Appendix_YesNo_Qwen2_5} respectively. On the online datasets, the model produces consistent outputs for some examples, while yielding different answers for others.

\section{Limitation}

As a preliminary exploration of using online data with verified reward to align LMMs, our work only study the alignment of image and text. This limits the ability of LMMs on other modality, such as video and audio. In the future, it is worth studying the effectiveness of using reinforcement learning with verified rewards to align LMMs on more modalities.

\end{document}